\documentclass{article}

\PassOptionsToPackage{numbers, compress}{natbib}

\usepackage[preprint]{neurips_2025}

\usepackage[utf8]{inputenc} 
\usepackage{xcolor}         
\usepackage[T1]{fontenc}    
\definecolor{cvprblue}{rgb}{0.21,0.49,0.74}
\usepackage[pagebackref,breaklinks,colorlinks,citecolor=cvprblue]{hyperref}   
\usepackage{url}            
\usepackage{booktabs}       
\usepackage{amsfonts}       
\usepackage{nicefrac}       
\usepackage{microtype}      
\usepackage{graphicx}
\usepackage{xspace}
\usepackage{amsmath} 
\usepackage{amssymb}        
\usepackage{etoc}
\usepackage{minitoc}
\usepackage{multirow}

\title{Virtual Fitting Room: Generating Arbitrarily Long Videos of Virtual Try-On from a Single Image\\Technical Preview}

\newcommand{\ourwork}{VFR\xspace}
\newcommand\themodel{\ourwork}

\newcommand{\yxw}[1]{} 
\newcommand{\jk}[1]{} 
\newcommand{\ab}[1]{} 
\newcommand{\mv}[1]{} 

\newcommand{\fig}[1]{{Fig$.$~\color{red}#1}}
\newcommand{\figs}[1]{{Figs$.$~\color{red}#1}}
\newcommand{\tab}[1]{{Tab$.$~\color{red}#1}}
\newcommand{\sect}[1]{{Sec$.$~\color{red}#1}}

\newcommand{\supplvid}[0]{{\href{https://immortalco.github.io/VirtualFittingRoom/}{project page}}}

\author{Jun-Kun Chen$^{1}$ \qquad Aayush Bansal$^{2}$ \qquad Minh Phuoc Vo$^{2}$\qquad Yu-Xiong Wang$^{1}$ \vspace{0.1em} \\ 
    $^1$University of Illinois Urbana-Champaign \qquad $^2$SpreeAI\vspace{0.1em}\\
    {\tt \hspace{0mm}\{junkun3,yxw\}@illinois.edu \qquad \tt \{aayush.bansal,minh.vo\}@spreeai.com} 
    \vspace{0.2em}\\
    {\tt \href{https://immortalco.github.io/VirtualFittingRoom/}{immortalco.github.io/VirtualFittingRoom}}
}

\begin{document}

\maketitle

\begin{abstract}

We introduce the Virtual Fitting Room (VFR), a novel video generative model that produces arbitrarily long virtual try-on videos. Our VFR models long video generation tasks as an auto-regressive, segment-by-segment generation process, eliminating the need for resource-intensive generation and lengthy video data, while providing the flexibility to generate videos of arbitrary length. The key challenges of this task are twofold: ensuring local smoothness between adjacent segments and maintaining global temporal consistency across different segments. To address these challenges, we propose our VFR framework, which ensures smoothness through a prefix video condition and enforces consistency with the anchor video—a 360$^{\circ}$ video that comprehensively captures the human's whole-body appearance. Our VFR generates minute-scale virtual try-on videos with both local smoothness and global temporal consistency under various motions, making it a pioneering work in long virtual try-on video generation.

\end{abstract}
\section{Introduction}
\label{sec:intro}

\begin{figure*}
  \includegraphics[width=\textwidth]{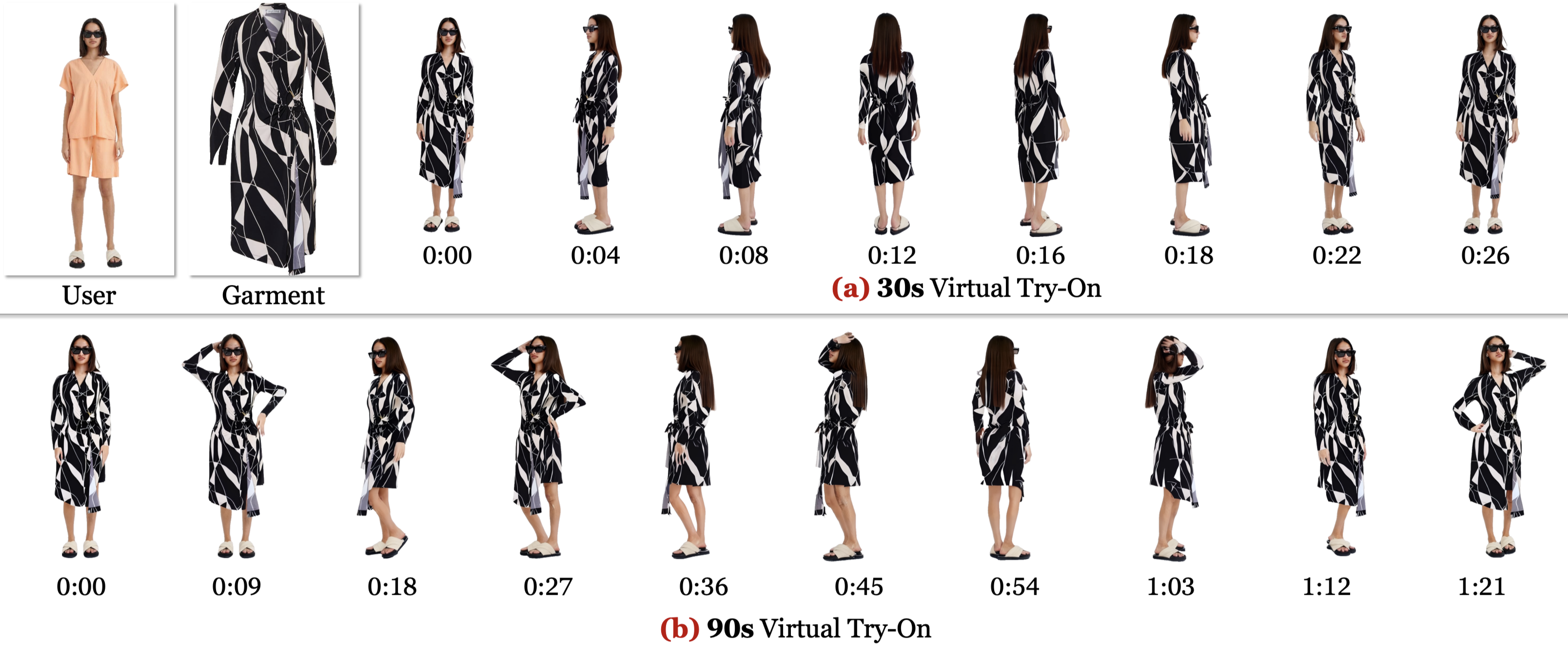}
  \vspace{-6mm}
  \caption{We generate two arbitrarily long videos: {\color{red}\bf (a)} a $30$s video; and {\color{red}\bf (b)} a $90$s video, for a user interacting with a given garment. Our approach preserves accessories -- glasses and slippers, and allows a desirable user-garment interaction. Please refer to the {\supplvid} for full streaming.}
  \label{fig:teaser}
  \vspace{-6mm}
\end{figure*}

Imagine being in a fitting room, trying on a garment, when a hurried knock interrupts you. Would that allow you to \emph{truly experience} the garment before buying it? No. To truly understand a garment, one may want to interact with it in various ways. The computational methods for \emph{virtually} trying on a garment enable a user to see themselves, but only in an image~\cite{4dvton,gpvton,tpd,stablevition,stablegarment,ootdiffusion} or a short 5$\sim$10s video~\cite{fashionvdm,dnd}, limiting the user's ability to fully experience a garment. We introduce \emph{Virtual Fitting Room} (\themodel) to enable a user to study the interaction of garments with their body \emph{as long as they like}. Unlike existing image or video try-on methods, \themodel allows a user to create \emph{arbitrarily} long videos ($720\times1152$ resolution at $8$ FPS and can be further refined to $24$ FPS) of themselves, given a single user image, a desired garment, and a reference video performing the desired try-on motion. \fig{\ref{fig:teaser}} shows a $30$s and a $90$s video generated using our method.

\begin{figure*}
  \includegraphics[width=\textwidth]{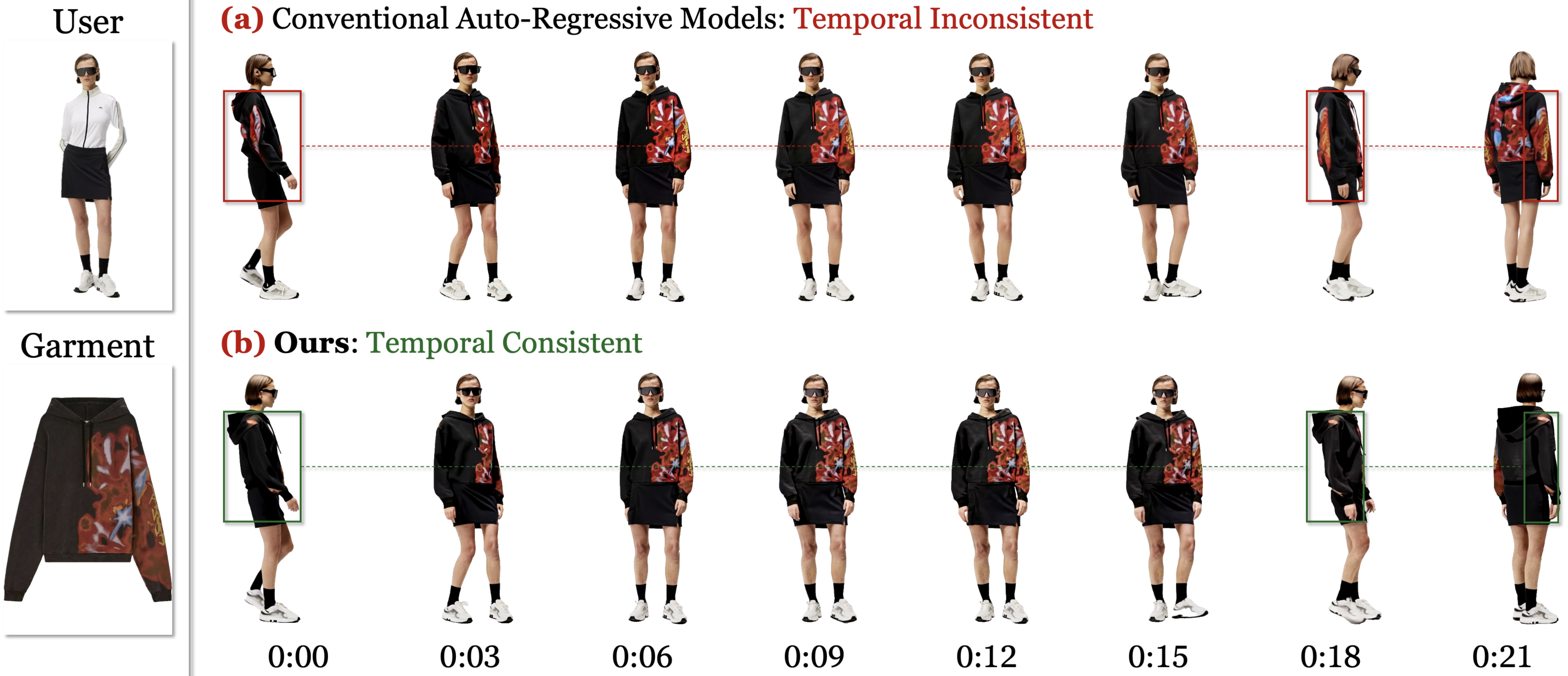}
  \vspace{-6mm}
  \caption{{Given a user and target garment, \bf\color{red}(a)} conventional auto-regressive video generators suffer from temporal inconsistency issues between distant frames. Note different patterns of sleeves in red bounding boxes across the time. {\bf\color{red}(b)} Our \themodel generates temporally consistent try-on videos.}
  \label{fig:temporal-consistent}
  \vspace{-2mm}
\end{figure*}

\begin{figure*}
  \includegraphics[width=\textwidth]{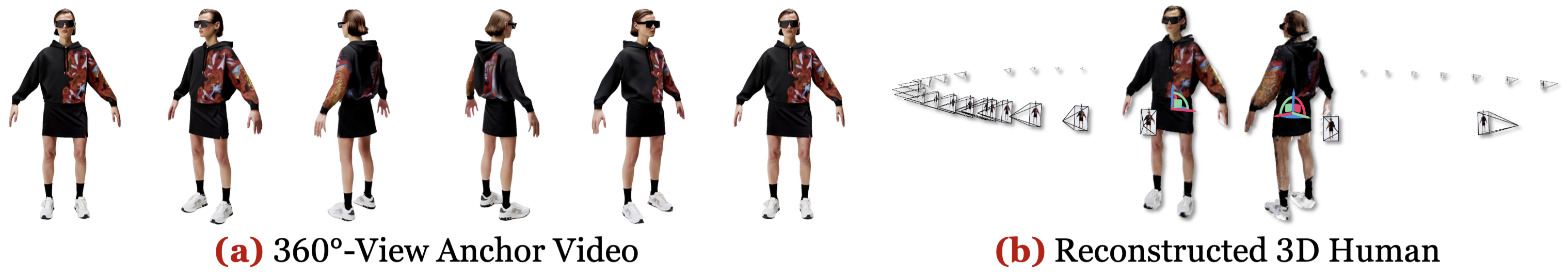}
  \vspace{-6mm}
  \caption{{\bf\color{red}(a)} VFR produces a 360$^{\circ}$ anchor video (``A'' pose) of a user for a given garment. We observe that the outputs are 3D-consistent, allowing us to {\bf\color{red}(b)} reconstruct it into a 3D human mesh.}
  \label{fig:3d}
  \vspace{-6mm}
\end{figure*}

\begin{figure*}
  \includegraphics[width=\textwidth]{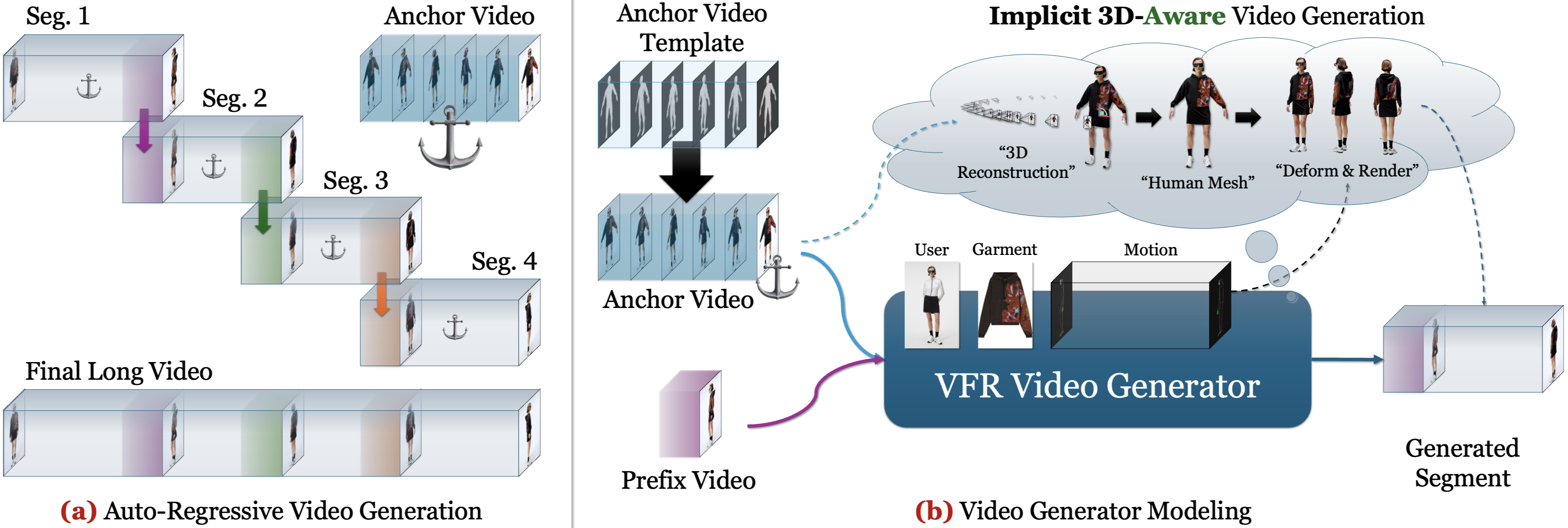}
  \vspace{-6mm}
  \caption{{\bf\color{red} (a)}  Our \themodel is an auto-regressive framework that generates a long video segment-by-segment. {\bf\color{red} (b)} The video generator model takes both an anchor video and a prefix video as input, and generates a new segment that continues the prefix video while maintaining consistency with the anchor video. }
  \label{fig:method}
  \vspace{-6mm}
\end{figure*}

Generating a 5s-long video is already a computationally demanding task~\cite{fashionvdm,dnd}. Naively extending these methods requires even more computational resources and a large-scale video dataset containing long videos for learning. To overcome these limitations, one may generate multiple short segments of a long video one by one ``auto-regressively'' in timestamp order, and then merge them to create a long video, as visualized in \fig{\ref{fig:method}-(a)}. Inspired by common approaches in general long video generation \cite{ttt,diffforce,historyguide,causvid,streamingt2v,framepack}, we allow each segment to slightly \emph{overlap} with the previous segment, and pass the overlapped ``prefix'' of the current segment as a condition, ensuring a \emph{locally} smooth transition between each adjacent segment pair.  However, these generated videos lack \emph{global} temporal consistency, as shown in \fig{\ref{fig:temporal-consistent}-(a)}, which is difficult to fix \emph{after} once they are generated. In this work, we draw inspiration from the process of writing an essay with an \emph{outline} as an ``anchor.'' We posit that creating long videos is a two-step process that involves: (1) creating an ``outline'' or an ``anchor'' to guide the generation; and (2) generating multiple short videos that are consistent with the anchor. We observe that a short 360$^{\circ}$ video like \fig{\ref{fig:3d}-(a)} of the human subject in a simple ``A'' pose serves as a reasonable anchor, allowing the model to comprehensively design the whole-body appearance of the human. The multiple short video segments are \emph{consistent} with the anchor video and, therefore, are also consistent with each other. With the anchor video, our generated long video achieves temporal consistency (\fig{\ref{fig:temporal-consistent}-(b)}) \emph{without} requiring long videos for training.

\textbf{Evaluating the quality of long video virtual try-on.} With the flexibility to generate arbitrarily long videos, we introduce an evaluation protocol to assess performance across four aspects with varying difficulty levels: 
(1) \textbf{360$^{\circ}$ Garment Consistency} -- a 360$^{\circ}$ video of a stationary human subject in ``A'' pose, which allows us to study the quality of the generated garment (\fig{\ref{fig:3d}-(a)}); 
(2) \textbf{360$^{\circ}$ Human+Garment Consistency} -- a 30s video of a human subject casually moving around a point in front of a stationary camera, which enables us to assess the quality of the generated human and garment (\fig{\ref{fig:teaser}-(a)}); 
(3) \textbf{Hand-Body Interaction Faithfulness} -- a 90s video of a human subject performing a fixed set of poses in front of a stationary camera, which facilitates the evaluation of the robustness of the virtual try-on method in controlled settings (\fig{\ref{fig:teaser}-(b)}); and 
(4) \textbf{Capability for Arbitrary Poses} - a 30$\sim$60s video of a human subject freely interacting with their body, which allows us to investigate robustness in various poses and orientations. 
We believe that this evaluation protocol will enable us to comprehensively assess the quality of virtual try-on methods.

\textbf{\emph{Free} viewpoint rendering is for \emph{free}.} A by-product of learning temporally consistent video is that we can render a human subject in any pose and viewpoint. \fig{\ref{fig:3d}-(a)} shows a generated 360$^{\circ}$ anchor video (``A'' pose) of a user wearing the target garment. The output is 3D consistent, enabling us to reconstruct it into a 3D mesh, as visualized in \fig{\ref{fig:3d}-(b)} in a NeRFStudio \cite{nerfstudio} viewer. 
Our observation indicates that 3D \emph{implicitly emerges} 

while enforcing temporal consistency. Interestingly, we can \emph{recloth} and \emph{remotion} any human subject from a single image, and view them from any viewpoint.

\textbf{Our Contributions.} (1) We introduce \themodel, a method to generate arbitrarily long, high-resolution ($720\times1152$
resolution at $24$FPS) human videos of virtual try-on from a single image. To our knowledge, no previous work has demonstrated these results. 
(2)  We also introduce an evaluation protocol to assess the overall quality of virtual try-on methods. Finally, (3) we observe that the proposed method implicitly learns 3D consistency, enabling us to perform free viewpoint rendering. 

\section{Related Work}
\label{sec:related}

We believe \emph{static 2D imagery isn't enough for a realistic virtual try-on experience}. The challenge of achieving high-resolution virtual try-on escalates as we progress from images to videos, and ultimately to 4D. This progression not only heightens the demand for computational resources, but also diminishes the availability of previous extensive databases necessary for learning. Our goal is to identify computational methods that enable the generation of arbitrarily long videos, even with constrained resources.

\noindent\textbf{Image-to-Image Try-on.}  Given an image of the user and a reference to the target garment, the goal here is to synthesize a new image of the user wearing the target garment~\cite{han2018viton,wang2018toward,issenhuth2020not,yang2020towards,lewis2021tryongan,ge2021parser,choi2021viton,morelli2022dress,bai2022single,dong2022dressing,he2022style,lee2022high,yang2022full,li2023virtual,morelli2023ladi,xie2023gp,yan2023linking,zhu2023tryondiffusion,10147324,yang2024texture,choi2024improving,zhang2024acdgvtonaccuratecontaineddiffusion,kim2024stableviton,li2024controlling,xu2025ootdiffusion,chongcatvton}. Most methods take a two-step approach: (1) deform the garment for a given user, known as warping; and (2) generate a new image with the deformed garment using GAN~\cite{goodfellow2014generative} or latent diffusion models~\cite{rombach2022high,podell2023sdxl,peebles2023scalable}. A few exceptions~\cite{lewis2021tryongan,zhu2023tryondiffusion,zhu2024m} generate the output without an intermediate warping step. The primary limitation in this field is the restricted experience a user can have with garments. One can only see them in exactly the same pose as in the input image. This issue can potentially be addressed by incorporating an additional module that can synthesize humans in different poses~\cite{ma2017pose,balakrishnan2018synthesizing,esser2018variational,siarohin2018deformable,men2020controllable,tang2020xinggan}. However, due to error propagation, the garment's appearance becomes inconsistent. Consequently, various methods~\cite{raj2018swapnet,han2019clothflow,sarkar2020neural,Cui_2021_ICCV,sarkar2021style} aim to jointly change the pose and the garment. These methods, however, lack temporal consistency and smoothness when applied across a series of poses. 

\noindent\textbf{Video-to-Video Try-on.} Given a video of the user and a reference to the target garment, the objective here is to synthesize a new video of the user wearing the target garment~\cite{kuppa2021shineon,zhong2021mv,jiang2022clothformer,xu2024tunnel,fang2024vivid,wang2024gpd}. An important distinction is ensuring that the synthesized garments and humans are temporally consistent and accurate. \citet{he2024wildvidfit} employs an image-to-image try-on methodology, but incorporates an additional temporal loss during training to enforce consistency. Recent methods~\cite{fashionvdm,dnd} can generate high-resolution output, but they are limited by the duration of the generated video ($5s$). Naively increasing the duration of videos will require enormous computational resources.

\noindent\textbf{Image-to-Video Try-on.} In this work, we explore the generation of arbitrarily long videos from a single user image and garment images. A naive approach is to utilize a single image try-on method and animate it using an image-to-video creation module~\cite{chan2019dance,Siarohin_2019_NeurIPS,Holynski_2021_CVPR,siarohin2021motion,xu2023magicanimate,guo2024animatediff,dreampose_2023,mimicmotion2024,azadi2023make,tevet2022human,guo2022generating,hu2024animate,zhu2024champ}. However, a modular approach leads to error propagation, which degrades the quality of the generated outputs. Therefore, we seek an end-to-end video generation pipeline that allows us to preserve the details of the garment~\cite{dnd, dong2019fw}. We observe that text conditioning cannot effectively capture long and subtle movements~\cite{dnd}. Instead, we use example videos as a reference to guide the creation of new videos. A notable prior work, DnD~\cite{dnd}, generates $5$s videos with high resolution $720\times1152$ at $24$FPS. In this work, we generate arbitrarily long videos from a single user image and the reference garments.

\noindent\textbf{Long Video Generation.} There is also a line of work \cite{ttt,diffforce,historyguide,causvid,streamingt2v,framepack} that investigates long video generation for text-to-video and image-to-video tasks. The common idea behind these methods is to introduce an additional ``memory back'' or ``history tracking'' mechanism to ensure consistency with the previous frames in a typical auto-regressive generation process. For example, a concurrent work, FramePack \cite{framepack} designs a computationally efficient way to consider all the previous frames as conditions when generating a new frame. These approaches can also generate \emph{smooth} long videos. However, their \emph{temporal consistency} is not guaranteed and often violated due to ineffective history tracking or over-compressed memory. Our \themodel method tackles this problem by designing the anchor video conditioning \emph{tailored for} virtual try-on tasks to promote temporal consistency across the long video.

\noindent\textbf{Free Viewpoint Rendering.} Our ability to generate arbitrarily long videos implicitly allows us to perform free-viewpoint rendering of humans~\cite{weng_humannerf_2022_cvpr,liu2021neural}. We observe that a model trained to capture consistent temporal characteristics implicitly learns 3D consistency in its outputs. This analysis could potentially pave the way for 4D try-ons in the future. \mv{This point is great. However, we might need to do a gausian splatting or such to show case that the 3D is indeed reconstructed.} \jk{Yes we can - Fig3b is already the reconstructed 3D scene captured from nerfstudio UI, not just a visualization; but may put more in suppl}

\section{\themodel: Methodology Overview}
\label{sec:method}

\begin{figure*}
  \includegraphics[width=\textwidth]{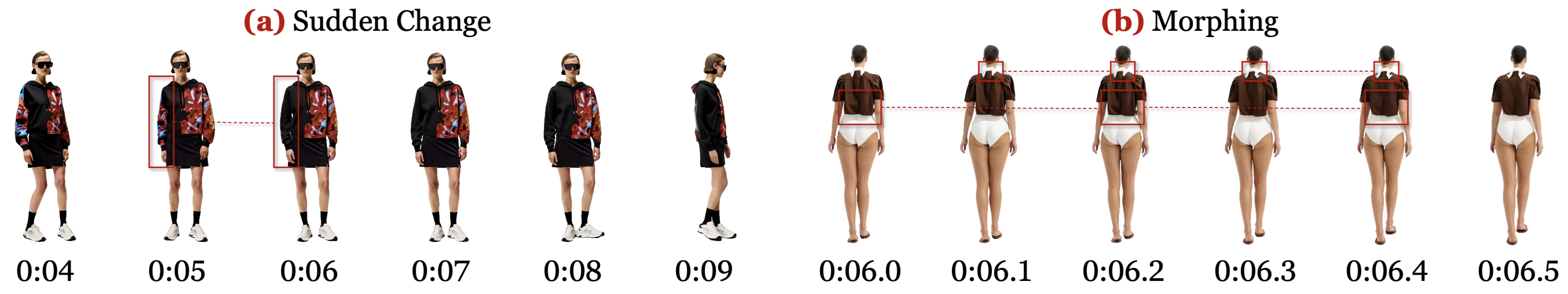}
  \vspace{-6mm}
  \caption{Without our prefix conditioning, the generated video may contain artifacts like {\color{red}\bf (a)} sudden changes or {\color{red}\bf (b)} morphing, as highlighted in the boxes, which violate the smoothness requirements. }
  \label{fig:unsmooth}
  \vspace{-6mm}
\end{figure*}

Given a user image, a reference garment image, an optional text prompt, and a long motion reference video, our \themodel is a method that generates long, minute-scale, high-quality try-on videos of the user wearing the desired garment and performing the indicated motion, in an auto-regressive, segment-by-segment manner. In an auto-regressive generation framework, the core challenge is to achieve both (1) \emph{local smoothness} such that the video transitions seamlessly without noticeable sudden changes or morphing (\fig{\ref{fig:unsmooth}}); and (2) \emph{global temporal consistency} such that the appearance of both the user and the garments at all occurrences in the video is the same (\fig{\ref{fig:temporal-consistent}}). 

To address these crucial challenges, our key insight is to (1) propose the ``anchor video'' generation to ensure a consistent appearance throughout the entire video, and (2) introduce the video prefix condition and immediate refiner to enhance video smoothness through strong conditioning. With both insights, our \themodel achieves high-quality long try-on videos, while ensuring both smoothness and temporal consistency. 

\section{Experiments}

\begin{figure*}[t!]
  \includegraphics[width=\textwidth]{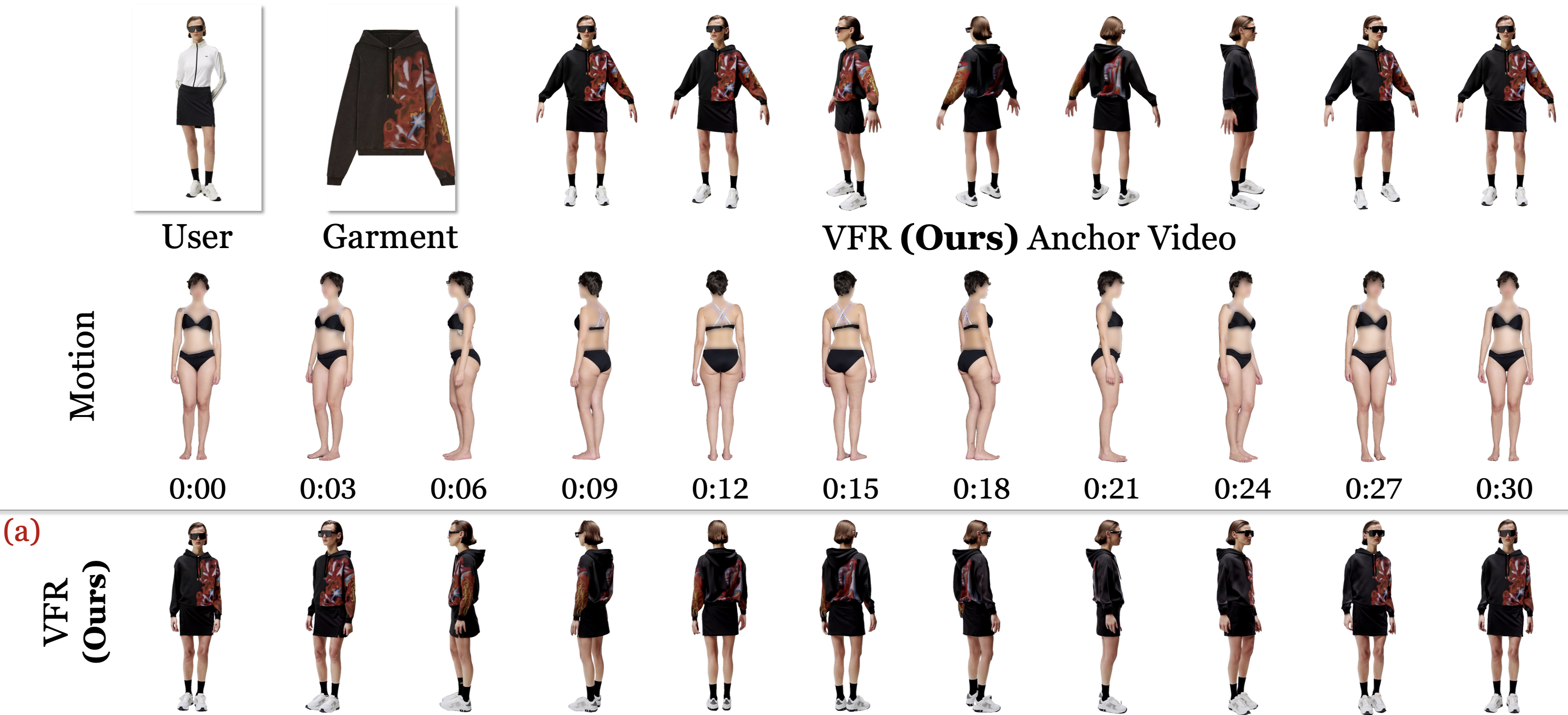}
  \includegraphics[width=\textwidth]{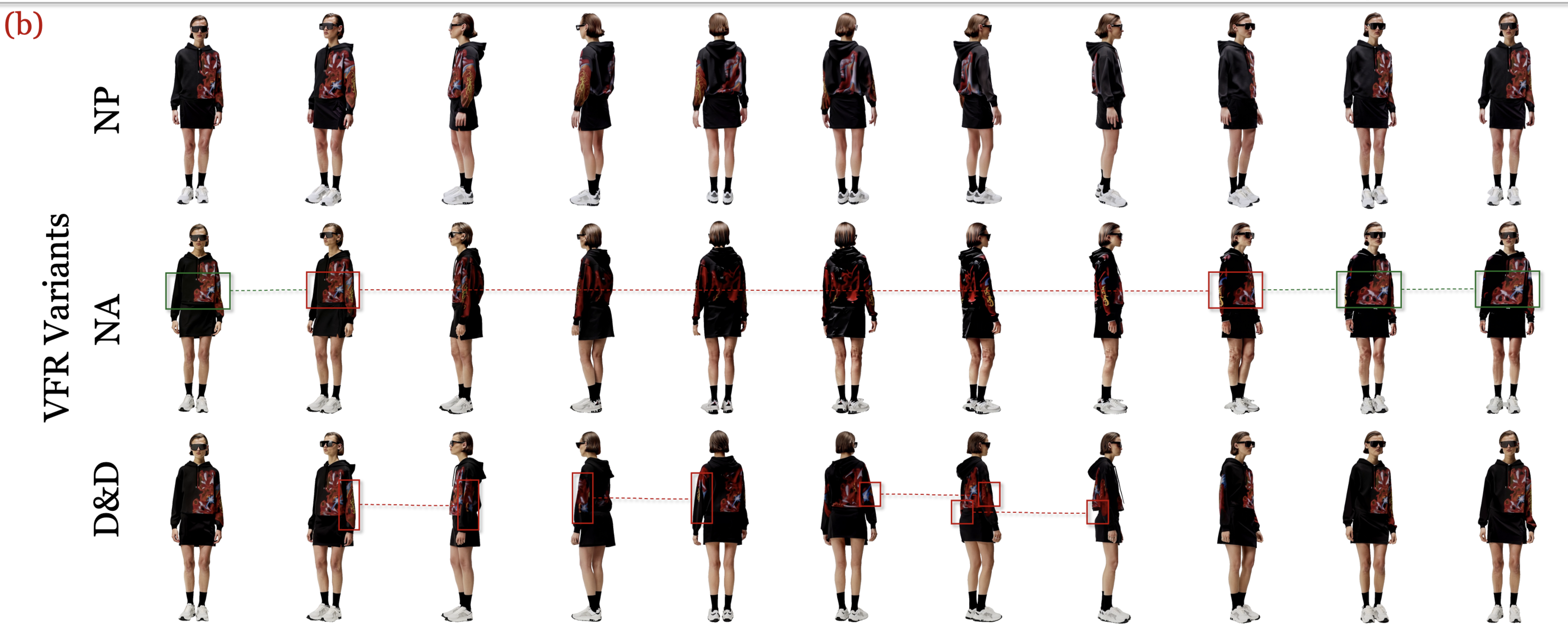}
  \includegraphics[width=\textwidth]{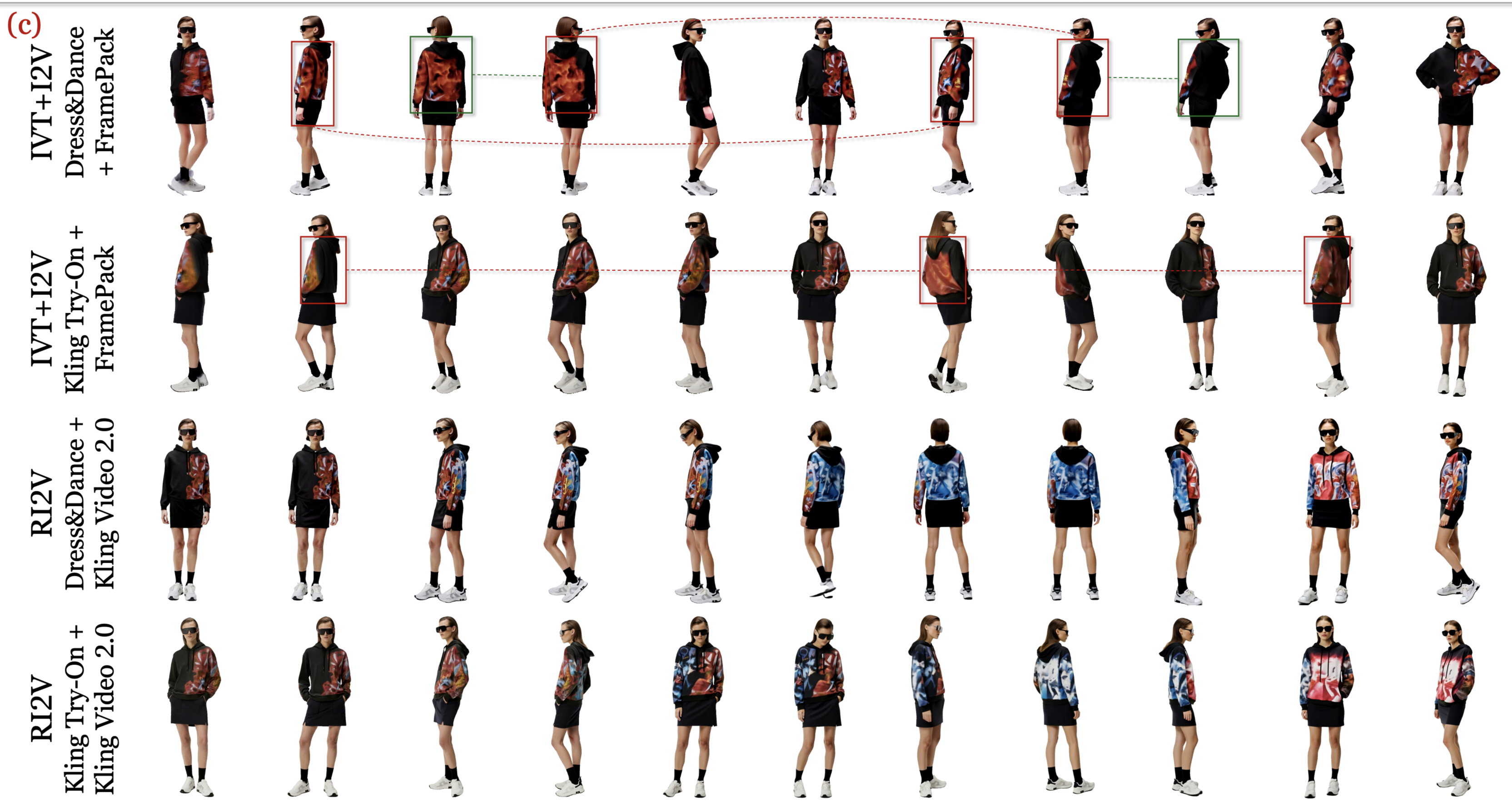}
  \vspace{-7mm}
  \caption{In the virtual try-on with 360$^{\circ}$ dynamic motion, {\color{red} (a)} our \themodel generates high quality, long virtual try-on videos, while {\color{red} (b)} removing either anchor video or prefix conditioning results in noticeable degradations. On the contrary, {\color{red} (c)} the baselines suffer from smoothness and temporal consistency issues. Please refer to the {\supplvid} for full streaming. }
  \label{fig:expr-360}
  \vspace{-6.5mm}
\end{figure*}

\subsection{Experimental Settings}

\noindent\textbf{Model Training Settings.} Our \themodel model is built on Dress\&Dance \cite{dnd} with the addition of ``prefix video'' and ``anchor video'' CondNets.  We train \themodel on both Internet and captured datasets from Dress\&Dance for $10,000$ iterations. Specifically, the immediate refiner is initialized from the $5,000$-th iteration checkpoint of the base \themodel, and trained for another $5,000$ iterations.

\noindent\textbf{Evaluation Tasks.} As mentioned in \sect{\ref{sec:intro}}, we have four different parts: 
(1) \textbf{360$^{\circ}$ Garment Consistency}, to generate a 5s 360$^{\circ}$-view ``A'' pose video, which also serves as the anchor videos for the other tasks; 
(2) \textbf{360$^{\circ}$ Human+Garment Consistency}, to generate a 30s 360$^{\circ}$ casually moving video;
(3) \textbf{360$^{\circ}$ Hand-Body Interaction Faithfulness}, to generate a 90s video with a fixed motion; and 
(4) \textbf{Capability for Arbitrary Poses}, to generate a 30$\sim$60s video with arbitrary motion.

\noindent\textbf{Baselines.} We compare our \themodel against the baselines \textbf{FramePack} \cite{framepack}, utilizing an ``image virtual try-on + image-to-video animation'' (IVT+I2V) procedure that is aligned with Dress\&Dance \cite{dnd}. We also compare with \textbf{Kling Video 2.0} \cite{kling} in a ``repeating image-to-video'' (RI2V) manner mentioned in \cite{framepack}. We are unable to compare with the following baselines: StreamT2V \cite{streamingt2v}, as it only supports 16:9 landscape videos; CausVid \cite{causvid}, given that there is no available image-to-video checkpoint released; TTT \cite{ttt}, since it only supports Tom and Jerry videos; and DiffusionForcing \cite{diffforce} and HistoryGuidance \cite{historyguide}, as they are restricted to videos from their respective trained datasets. As for the image try-on method, we mainly use the two state-of-the-art method, \textbf{Dress\&Dance} image try-on \cite{dnd} and \textbf{Kling Try-On} \cite{kling}, to generate the first frame of the video. 

\noindent\textbf{Ablation studies.} We also compare our full \themodel with the following variants:
(1) \textbf{``No Prefix'' (NP)}, which does not use prefix conditioning, but directly utilizes DiffEdit \cite{diffedit} to outpaint the video with the prefix;
(2) \textbf{``No Anchor'' (NA)}, which does not generate and condition the segment generations on the anchor video; 
(3) \textbf{``Dress\&Dance'' (D\&D)}, which does not use either the anchor video or the prefix conditioning, making it equivalent to a training-free method that employs Dress\&Dance with DiffEdit for long video generation.
(4) \textbf{``No Refine'' (NR)}, which does not use of the immediate refiner to refine each segment's output.

\noindent\textbf{Metrics.} Consistent with Dress\&Dance \cite{dnd} and FramePack \cite{framepack}, we utilize GPT \cite{gpt}-based scores and VBench \cite{vbench,vbench++,vbench2} to evaluate our videos. GPT scores can effectively assess the try-on quality from various aspects, leveraging GPT's visual capabilities; VBench introduces a set of metrics that comprehensively evaluate the videos from both quality and semantic perspectives.

\begin{figure*}[t!]
  \includegraphics[width=\textwidth]{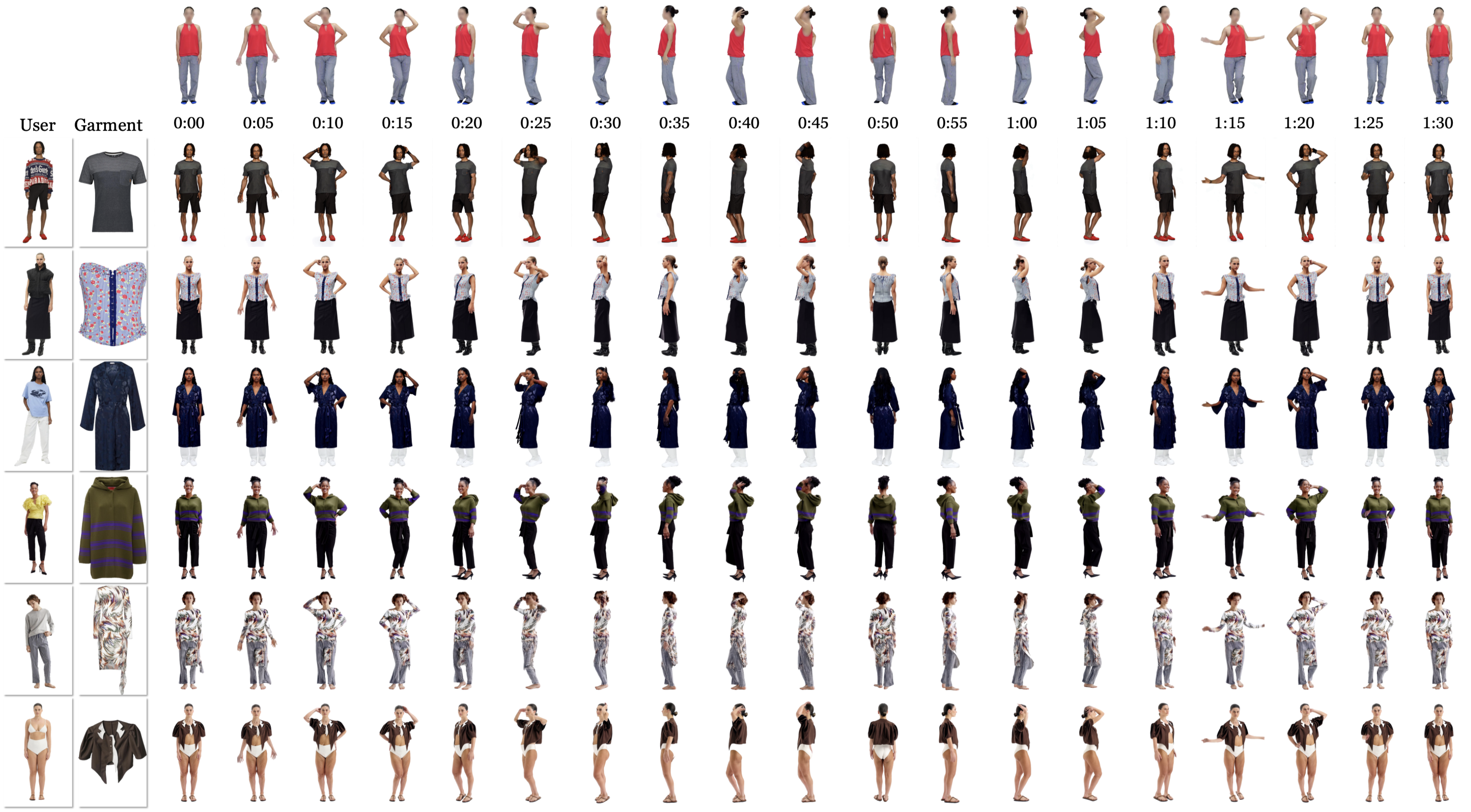}
  \vspace{-6mm}
  \caption{In the virtual try-on with the 90s hand-body interaction motion, our \themodel generates temporally consistent try-on videos. Please refer to the {\supplvid} for full streaming.}
  \label{fig:expr-template90}
  \vspace{-6mm}
\end{figure*}

\begin{figure*}[t!]
  \includegraphics[width=\textwidth]{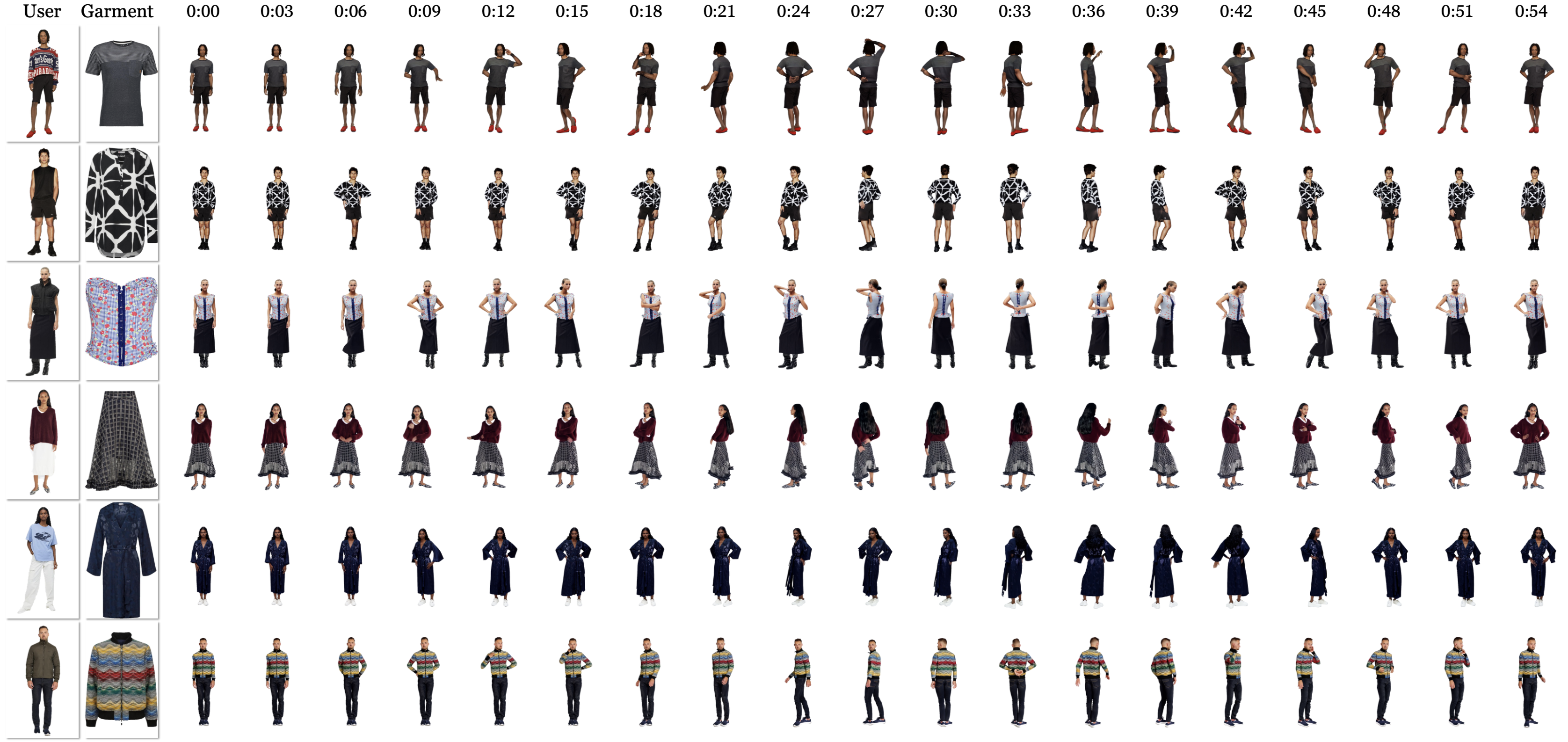}
  \vspace{-6mm}
  \caption{In the virtual try-on with a $\sim$50s arbitrary motion, our \themodel faithfully preserves consistent garment details and human appearance, showcasing various poses with high quality. These results are shown as videos in our \supplvid.}
  \label{fig:expr-free50}
  \vspace{-6mm}
\end{figure*}

\subsection{Experimental Results and Analysis}

We present our qualitative results in \figs{\ref{fig:expr-360},\ref{fig:expr-template90},\ref{fig:expr-free50}} as images, and in our \supplvid{} as videos. We provide the quantitative results in \tab{\ref{tab:expr-quan}}.

\noindent\textbf{360$^{\circ}$ Human+Garment Consistency (30s) -- \fig{\ref{fig:expr-360}}.} As shown \fig{\ref{fig:expr-360}-(a)}, our \themodel produces high-quality, long virtual try-on videos.  In \fig{\ref{fig:expr-360}-(b)}, our ``No Prefix'' (NP) variant, due to the global anchor videos, generates results that are comparable to our full VFR, but exhibits some sudden changes, as illustrated in our \supplvid. In contrast, our ``No Anchor'' (NA) variant's video displays long-term temporal inconsistencies, while our ``Dress\&Dance training-free'' (D\&D) variant exhibits even greater temporal inconsistencies in both the short and long term. This highlights that both designs in our \themodel for local smoothness and global temporal consistency are effective and essential. In \fig{\ref{fig:expr-360}-(c)}, the baseline FramePack \cite{framepack} produces overall smooth results with long-term inconsistencies, while the appearances deviate significantly in the results produced by the Kling Video 2.0 \cite{kling}-based RI2V method. This demonstrates that the virtual try-on tasks are non-trivial and challenging, underscoring our contribution in achieving high-quality results.

\noindent\textbf{Hand-Body Interaction Faithfullness (90s) -- \fig{\ref{fig:expr-template90}}.} The motion in these evaluation tasks encompasses both human rotation and arm movements. Our \themodel faithfully performs the same motion in different virtual try-on tasks, demonstrating the ability to depict the same motion across various garments -- either pants, skirts, or dresses. Even for such a long-term video, our \themodel still maintains high temporal consistency, which can be observed by comparing the first and the last frame. 

\noindent\textbf{Capability for Arbitrary Poses ($\sim$50s) -- \fig{\ref{fig:expr-free50}}.} In these highly challenging tasks, the motions can be arbitrary, encompassing various arm and leg movements, which lead to even more diverse showcases and interactions between garments and users. We observe that our \themodel effectively handles these motions and generates high-quality visualizations to depict them. This shows \themodel has the capability to generalize to various long video virtual try-on tasks.

\begin{table*}[t!]
\centering
\renewcommand{\arraystretch}{1.2}
\setlength{\tabcolsep}{3pt}
\scalebox{0.55}{
\begin{tabular}{ll|ccc|ccccc}
\hline\hline
\multicolumn{2}{c|}{\textbf{Method}} & Subject Consistency & Background Consistency & Motion Smoothness & $\text{GPT}_\text{Try-On}\uparrow$ &  $\text{GPT}_\text{User}\uparrow$  & $\text{GPT}_\text{Motion}\uparrow$ & $\text{GPT}_\text{Visual}\uparrow$ & $\text{GPT}_\text{Overall}\uparrow$ \\

\hline
\multicolumn{10}{c}{\bf 1. 360$^{\circ}$ Garment Consistency (5s)}\\
\hline 
Ours & & 92.84 & 95.38 & 98.35& 87.83 & 85.90 & 76.67 & 80.35 & 82.06\\
\hline
\multicolumn{10}{c}{\bf 2. 360$^{\circ}$ Human+Garment Consistency (30s)}\\
\hline 
Ours & & \textbf{94.06} & \textbf{96.53} & \textbf{99.37}& \textbf{90.09} & \textbf{88.11} & 84.08 & \textbf{86.33} & \textbf{87.14} \\
Ours & NP & 93.58 & 96.10 & 99.31& 89.66 & 87.24 & \textbf{84.20} & 85.86 & 86.69 \\
Ours & NA & 92.77 & 95.55 & 99.22& 88.47 & 84.91 & 81.75 & 81.66 & 84.04 \\
Ours & D\&D & 93.70 & 96.01 & 99.23& 89.72 & 86.16 & 83.20 & 86.05 & 86.10 \\
Ours & NR & 90.80 & 94.82 & 99.21& 87.29 & 85.55 & 78.74 & 71.90 & 80.40 \\
\hline
\multicolumn{10}{c}{\bf 3. Hand-Body Interaction Faithfulness (90s)}\\
\hline 
Ours &  & \textbf{92.13} & \textbf{94.02} & \textbf{99.24}& 87.20 & 84.88 & 77.85 & 81.12 & 82.62 \\
Ours & NP & 91.88 & 93.91 & 99.15& 89.65 & 86.28 & 80.15 & 83.22 & 84.62 \\
Ours & NA & 88.00 & 91.50 & 99.09 & 85.20 & 80.35 & 70.67 & 69.05 & 75.70 \\
Ours & D\&D & 91.65 & 93.25 & 99.11 & \textbf{89.95} & \textbf{87.25} & \textbf{82.53} & \textbf{85.15} & \textbf{86.15} \\
Ours & NR & 86.02 & 89.84 & 99.01& 85.22 & 84.83 & 72.25 & 65.10 & 75.03 \\
\hline
\multicolumn{10}{c}{\bf 4-Med. Capability for Arbitrary Poses -- Medium (25s)}\\
\hline 
Ours&           & \textbf{91.09} & \textbf{93.82} & \textbf{98.62} & 87.11 & 84.45 & 77.83 & 79.77 & 81.93 \\
Ours&NP        & 90.19 & 93.00 & 97.85& 87.40 & \textbf{85.48} & 78.40 & 80.55 & 82.56 \\
Ours&NA        & 88.46 & 92.10 & 97.92& 86.87 & 84.39 & 77.92 & 77.56 & 81.31 \\
Ours&D\&D    & 89.27 & 92.81 & 97.83& \textbf{87.83} & 85.23 & \textbf{80.24} & \textbf{82.22} & \textbf{83.61} \\
Ours&NR        & 87.56 & 91.79 & 97.76& 85.28 & 83.75 & 74.85 & 68.98 & 77.26 \\
\hline 
\multicolumn{10}{c}{\bf 4-Hard. Capability for Arbitrary Poses -- Hard (30-50s)}\\
\hline 
Ours& & \textbf{93.21} & \textbf{95.29} & \textbf{99.35} & 87.03 & 85.83 & 69.99 & 78.84 & 79.05 \\
Ours&NP & 92.71 & 94.63 & 99.29 & 87.69 &\textbf{ 86.38} & 71.12 & 79.47 & 79.60 \\
Ours&NA & 91.17 & 93.31 & 99.22& 87.14 & 85.03 & 71.42 & 77.11 & 78.99 \\
Ours&D\&D & 92.27 & 94.09 & 99.25& \textbf{88.05} & 84.97 & 71.90 & \textbf{81.38} & 80.39 \\
Ours&NR & 88.17 & 92.08 & 99.19& 85.36 & 84.61 & 67.45 & 64.55 & 73.44 \\
IVT+I2V& D\&D + FramePack \cite{framepack} & 88.33 & 90.98 & 98.24& 86.96 & 85.35 & \textbf{77.51} & 79.33 & \textbf{81.86} \\
RI2V& D\&D + Kling \cite{kling} & 92.71 & 94.08 & 98.92& 87.25 & 85.59 & 65.80 & 79.40 & 77.24 \\
\hline

\hline\hline
\end{tabular}
}

\caption{Quantitative experiments show that our \themodel consistently outperforms all variants and baselines in both consistency and smoothness metrics from VBench \cite{vbench}, while achieving comparable try-on and visual quality as Dress\&Dance \cite{dnd}.}
\label{tab:expr-quan}

\end{table*}

\noindent\textbf{Quantitative Experiments.} The quantitative evaluation comparisons are provided in Table {\ref{tab:expr-quan}}. We use the ``Subject Consistency,'' ``Background Consistency,'' and ``Motion Smoothness'' from VBench \cite{vbench} to evaluate how the two major challenges -- temporal consistency and smoothness -- are addressed, as well as the GPT metric in \cite{dnd} to assess the overall virtual try-on quality.

As shown in Table {\ref{tab:expr-quan}}, in each component of the evaluation protocols, our full \themodel consistently outperforms all the baselines and variants. Furthermore, since our \themodel is based on the previous work Dress\&Dance \cite{dnd}, all these variants maintain a portion of its virtual try-on capability, achieving comparable GPT evaluation scores, where the D\&D variant, as a training-free long video generation pipeline utilizing D\&D, preserves the majority of its capability. 

We further observe that our NP variant achieves performance very similar to that of our full model, reflecting the strong control provided by the anchor video. Without the anchor video, however, the NA and D\&D variants suffer a significant drop in both (human) subject and background consistencies, and the degradation of virtual try-on quality even occurs for the NA variant, as reflected in the GPT evaluation metric. This shows that the anchor video controls not only the consistent appearance but also the try-on quality.

Finally, we compare the most powerful baselines, FramePack \cite{framepack} and Kling \cite{kling}, combined with the state-of-the-art virtual try-on method Dress\&Dance. FramePack has a significantly lower consistency metric, indicating that its conditioning method, which takes into account the previous frames, is not sufficient to enforce consistency. Kling still achieves a slightly lower consistency metric compared to our \themodel, but the quality of the virtual try-on degrades.

\section{Conclusion}
\label{sec:conclusion}

We propose \themodel, a virtual try-on method that generates arbitrarily long, high-resolution videos from a single user image, garment, and motion reference. The key of our method is an anchor video-guided framework that ensures temporal consistency across segments and implicitly captures 3D structure, enabling free-viewpoint rendering without 3D supervision. Along with the prefix conditioning, we achieve both local smoothness and global temporal consistency in the long video generation.  We further introduce a new evaluation protocol tailored to long video virtual try-on, covering garment fidelity, user appearance, and motion robustness. Experiments show that \themodel produces realistic, smooth, temporally consistent, and garment-faithful results that significantly surpass the capabilities of prior methods. We believe that \themodel opens up new avenues for interactive, personalized virtual try-on experiences—whether in e-commerce, virtual social platforms, or creative content generation. 

\noindent\textbf{Discussions.} We made progress in generating arbitrarily long, high-resolution videos for virtual try-on. Our preliminary analysis shows that additional video data can help us further improve the quality of the generated videos. Secondly, while this work is the first of its kind, it takes 1$\sim$2 hours to generate a $30$s video, which is not efficient enough to produce long videos in nearly real-time -- we leave this speed-up as an interesting future work. Finally, we believe that our work will pave the way for the transition from long videos to arbitrary 4D content, where a user can both change camera perspective and motion.

\noindent\textbf{Potential Societal Impacts.} The positive societal impacts of our \themodel may include (1) revolutionizing the online shopping experience for clothing, (2) decreasing returns and replacements of clothes through improved pre-sale understanding, and (3) leading to an increase in both the number and volume of online clothing shops. On the other hand, \themodel is inherently a model that produces human videos, and also brings the risks to produce biased, unethical, or unsafe results.

\newpage
{
    \small
    \bibliographystyle{unsrtnat}
    \bibliography{main}

\begin{thebibliography}{86}
\providecommand{\natexlab}[1]{#1}
\providecommand{\url}[1]{\texttt{#1}}
\expandafter\ifx\csname urlstyle\endcsname\relax
  \providecommand{\doi}[1]{doi: #1}\else
  \providecommand{\doi}{doi: \begingroup \urlstyle{rm}\Url}\fi

\bibitem[Yang et~al.(2024{\natexlab{a}})Yang, Jiang, Li, Zhou, Dong, Zhang, and Du]{4dvton}
Zhaotong Yang, Zicheng Jiang, Xinzhe Li, Huiyu Zhou, Junyu Dong, Huaidong Zhang, and Yong Du.
\newblock D 4-vton: Dynamic semantics disentangling for differential diffusion based virtual try-on.
\newblock In \emph{European Conference on Computer Vision}, pages 36--52. Springer, 2024{\natexlab{a}}.

\bibitem[Xiel et~al.(2023)Xiel, Huang, Dong, Zhao, Dong, Zhang, Zhu, and Liang]{gpvton}
Zhenyu Xiel, Zaiyu Huang, Xin Dong, Fuwei Zhao, Haoye Dong, Xijin Zhang, Feida Zhu, and Xiaodan Liang.
\newblock Gp-vton: Towards general purpose virtual try-on via collaborative local-flow global-parsing learning.
\newblock In \emph{ArXiV}, pages 23550--23559, 06 2023.
\newblock \doi{10.1109/CVPR52729.2023.02255}.

\bibitem[Yang et~al.(2024{\natexlab{b}})Yang, Ding, Hong, Huang, Tao, and Xu]{tpd}
Xu~Yang, Changxing Ding, Zhibin Hong, Junhao Huang, Jin Tao, and Xiangmin Xu.
\newblock Texture-preserving diffusion models for high-fidelity virtual try-on, 2024{\natexlab{b}}.

\bibitem[Kim et~al.(2024{\natexlab{a}})Kim, Gu, Park, Park, and Choo]{stablevition}
Jeongho Kim, Gyojung Gu, Minho Park, Sunghyun Park, and Jaegul Choo.
\newblock Stableviton: Learning semantic correspondence with latent diffusion model for virtual try-on.
\newblock In \emph{CVPR}, 2024{\natexlab{a}}.

\bibitem[Wang et~al.(2024{\natexlab{a}})Wang, Guo, Liu, Li, Zhao, Tang, Hu, Tang, and Li]{stablegarment}
Rui Wang, Hailong Guo, Jiaming Liu, Huaxia Li, Haibo Zhao, Xu~Tang, Yao Hu, Hao Tang, and Peipei Li.
\newblock Stablegarment: Garment-centric generation via stable diffusion.
\newblock \emph{arXiv preprint arXiv:2403.10783}, 2024{\natexlab{a}}.

\bibitem[Xu et~al.(2024{\natexlab{a}})Xu, Gu, Chen, and Chen]{ootdiffusion}
Yuhao Xu, Tao Gu, Weifeng Chen, and Chengcai Chen.
\newblock Ootdiffusion: Outfitting fusion based latent diffusion for controllable virtual try-on.
\newblock \emph{arXiv preprint arXiv:2403.01779}, 2024{\natexlab{a}}.

\bibitem[Karras et~al.(2024)Karras, Li, Liu, Zhu, Yoo, Lugmayr, Lee, and Kemelmacher-Shlizerman]{fashionvdm}
Johanna Karras, Yingwei Li, Nan Liu, Luyang Zhu, Innfarn Yoo, Andreas Lugmayr, Chris Lee, and Ira Kemelmacher-Shlizerman.
\newblock Fashion-vdm: Video diffusion model for virtual try-on.
\newblock In \emph{Proceedings of ACM SIGGRAPH Asia 2024}, December 2024.

\bibitem[Author(s)(2025)]{dnd}
Anonymous Author(s).
\newblock {Dress\&Dance}: Dress up and dance as you like it.
\newblock In \emph{Under Review}, January 2025.
\newblock URL \url{https://anonymous.4open.science/r/Dress_and_Dance_paper}.

\bibitem[Dalal et~al.(2025)Dalal, Koceja, Hussein, Xu, Zhao, Song, Han, Cheung, Kautz, Guestrin, Hashimoto, Koyejo, Choi, Sun, and Wang]{ttt}
Karan Dalal, Daniel Koceja, Gashon Hussein, Jiarui Xu, Yue Zhao, Youjin Song, Shihao Han, Ka~Chun Cheung, Jan Kautz, Carlos Guestrin, Tatsunori Hashimoto, Sanmi Koyejo, Yejin Choi, Yu~Sun, and Xiaolong Wang.
\newblock One-minute video generation with test-time training, 2025.
\newblock URL \url{https://arxiv.org/abs/2504.05298}.

\bibitem[Chen et~al.(2025)Chen, Mart{\'\i}~Mons{\'o}, Du, Simchowitz, Tedrake, and Sitzmann]{diffforce}
Boyuan Chen, Diego Mart{\'\i}~Mons{\'o}, Yilun Du, Max Simchowitz, Russ Tedrake, and Vincent Sitzmann.
\newblock Diffusion forcing: Next-token prediction meets full-sequence diffusion.
\newblock \emph{Advances in Neural Information Processing Systems}, 37:\penalty0 24081--24125, 2025.

\bibitem[Song et~al.(2025)Song, Chen, Simchowitz, Du, Tedrake, and Sitzmann]{historyguide}
Kiwhan Song, Boyuan Chen, Max Simchowitz, Yilun Du, Russ Tedrake, and Vincent Sitzmann.
\newblock History-guided video diffusion, 2025.
\newblock URL \url{https://arxiv.org/abs/2502.06764}.

\bibitem[Yin et~al.(2025)Yin, Zhang, Zhang, Freeman, Durand, Shechtman, and Huang]{causvid}
Tianwei Yin, Qiang Zhang, Richard Zhang, William~T Freeman, Fredo Durand, Eli Shechtman, and Xun Huang.
\newblock From slow bidirectional to fast autoregressive video diffusion models.
\newblock In \emph{CVPR}, 2025.

\bibitem[Henschel et~al.(2024)Henschel, Khachatryan, Hayrapetyan, Poghosyan, Tadevosyan, Wang, Navasardyan, and Shi]{streamingt2v}
Roberto Henschel, Levon Khachatryan, Daniil Hayrapetyan, Hayk Poghosyan, Vahram Tadevosyan, Zhangyang Wang, Shant Navasardyan, and Humphrey Shi.
\newblock Streamingt2v: Consistent, dynamic, and extendable long video generation from text.
\newblock \emph{arXiv preprint arXiv:2403.14773}, 2024.

\bibitem[Zhang and Agrawala(2025)]{framepack}
Lvmin Zhang and Maneesh Agrawala.
\newblock Packing input frame contexts in next-frame prediction models for video generation.
\newblock \emph{Arxiv}, 2025.

\bibitem[Tancik et~al.(2023)Tancik, Weber, Ng, Li, Yi, Kerr, Wang, Kristoffersen, Austin, Salahi, Ahuja, McAllister, and Kanazawa]{nerfstudio}
Matthew Tancik, Ethan Weber, Evonne Ng, Ruilong Li, Brent Yi, Justin Kerr, Terrance Wang, Alexander Kristoffersen, Jake Austin, Kamyar Salahi, Abhik Ahuja, David McAllister, and Angjoo Kanazawa.
\newblock {Nerfstudio}: A modular framework for neural radiance field development.
\newblock In \emph{SIGGRAPH}, 2023.

\bibitem[Han et~al.(2018)Han, Wu, Wu, Yu, and Davis]{han2018viton}
Xintong Han, Zuxuan Wu, Zhe Wu, Ruichi Yu, and Larry~S Davis.
\newblock Viton: An image-based virtual try-on network.
\newblock In \emph{Proceedings of the IEEE conference on computer vision and pattern recognition}, pages 7543--7552, 2018.

\bibitem[Wang et~al.(2018)Wang, Zheng, Liang, Chen, Lin, and Yang]{wang2018toward}
Bochao Wang, Huabin Zheng, Xiaodan Liang, Yimin Chen, Liang Lin, and Meng Yang.
\newblock Toward characteristic-preserving image-based virtual try-on network.
\newblock In \emph{Proceedings of the European conference on computer vision (ECCV)}, pages 589--604, 2018.

\bibitem[Issenhuth et~al.(2020)Issenhuth, Mary, and Calauzenes]{issenhuth2020not}
Thibaut Issenhuth, J{\'e}r{\'e}mie Mary, and Cl{\'e}ment Calauzenes.
\newblock Do not mask what you do not need to mask: a parser-free virtual try-on.
\newblock In \emph{Computer Vision--ECCV 2020: 16th European Conference, Glasgow, UK, August 23--28, 2020, Proceedings, Part XX 16}, pages 619--635. Springer, 2020.

\bibitem[Yang et~al.(2020)Yang, Zhang, Guo, Liu, Zuo, and Luo]{yang2020towards}
Han Yang, Ruimao Zhang, Xiaobao Guo, Wei Liu, Wangmeng Zuo, and Ping Luo.
\newblock Towards photo-realistic virtual try-on by adaptively generating-preserving image content.
\newblock In \emph{Proceedings of the IEEE/CVF conference on computer vision and pattern recognition}, pages 7850--7859, 2020.

\bibitem[Lewis et~al.(2021)Lewis, Varadharajan, and Kemelmacher-Shlizerman]{lewis2021tryongan}
Kathleen~M Lewis, Srivatsan Varadharajan, and Ira Kemelmacher-Shlizerman.
\newblock Tryongan: Body-aware try-on via layered interpolation.
\newblock \emph{ACM Transactions on Graphics (TOG)}, 40\penalty0 (4):\penalty0 1--10, 2021.

\bibitem[Ge et~al.(2021)Ge, Song, Zhang, Ge, Liu, and Luo]{ge2021parser}
Yuying Ge, Yibing Song, Ruimao Zhang, Chongjian Ge, Wei Liu, and Ping Luo.
\newblock Parser-free virtual try-on via distilling appearance flows.
\newblock In \emph{Proceedings of the IEEE/CVF conference on computer vision and pattern recognition}, pages 8485--8493, 2021.

\bibitem[Choi et~al.(2021)Choi, Park, Lee, and Choo]{choi2021viton}
Seunghwan Choi, Sunghyun Park, Minsoo Lee, and Jaegul Choo.
\newblock Viton-hd: High-resolution virtual try-on via misalignment-aware normalization.
\newblock In \emph{Proceedings of the IEEE/CVF conference on computer vision and pattern recognition}, pages 14131--14140, 2021.

\bibitem[Morelli et~al.(2022)Morelli, Fincato, Cornia, Landi, Cesari, and Cucchiara]{morelli2022dress}
Davide Morelli, Matteo Fincato, Marcella Cornia, Federico Landi, Fabio Cesari, and Rita Cucchiara.
\newblock Dress code: High-resolution multi-category virtual try-on.
\newblock In \emph{Proceedings of the IEEE/CVF conference on computer vision and pattern recognition}, pages 2231--2235, 2022.

\bibitem[Bai et~al.(2022)Bai, Zhou, Li, Zhou, and Yang]{bai2022single}
Shuai Bai, Huiling Zhou, Zhikang Li, Chang Zhou, and Hongxia Yang.
\newblock Single stage virtual try-on via deformable attention flows.
\newblock In \emph{European Conference on Computer Vision}, pages 409--425. Springer, 2022.

\bibitem[Dong et~al.(2022)Dong, Zhao, Xie, Zhang, Du, Zheng, Long, Liang, and Yang]{dong2022dressing}
Xin Dong, Fuwei Zhao, Zhenyu Xie, Xijin Zhang, Daniel~K Du, Min Zheng, Xiang Long, Xiaodan Liang, and Jianchao Yang.
\newblock Dressing in the wild by watching dance videos.
\newblock In \emph{Proceedings of the IEEE/CVF Conference on Computer Vision and Pattern Recognition}, pages 3480--3489, 2022.

\bibitem[He et~al.(2022)He, Song, and Xiang]{he2022style}
Sen He, Yi-Zhe Song, and Tao Xiang.
\newblock Style-based global appearance flow for virtual try-on.
\newblock In \emph{Proceedings of the IEEE/CVF conference on computer vision and pattern recognition}, pages 3470--3479, 2022.

\bibitem[Lee et~al.(2022)Lee, Gu, Park, Choi, and Choo]{lee2022high}
Sangyun Lee, Gyojung Gu, Sunghyun Park, Seunghwan Choi, and Jaegul Choo.
\newblock High-resolution virtual try-on with misalignment and occlusion-handled conditions.
\newblock In \emph{European Conference on Computer Vision}, pages 204--219. Springer, 2022.

\bibitem[Yang et~al.(2022)Yang, Yu, and Liu]{yang2022full}
Han Yang, Xinrui Yu, and Ziwei Liu.
\newblock Full-range virtual try-on with recurrent tri-level transform.
\newblock In \emph{Proceedings of the IEEE/CVF conference on computer vision and pattern recognition}, pages 3460--3469, 2022.

\bibitem[Li et~al.(2023{\natexlab{a}})Li, Wei, Yin, Ma, and Kot]{li2023virtual}
Zhi Li, Pengfei Wei, Xiang Yin, Zejun Ma, and Alex~C Kot.
\newblock Virtual try-on with pose-garment keypoints guided inpainting.
\newblock In \emph{Proceedings of the IEEE/CVF international conference on computer vision}, pages 22788--22797, 2023{\natexlab{a}}.

\bibitem[Morelli et~al.(2023)Morelli, Baldrati, Cartella, Cornia, Bertini, and Cucchiara]{morelli2023ladi}
Davide Morelli, Alberto Baldrati, Giuseppe Cartella, Marcella Cornia, Marco Bertini, and Rita Cucchiara.
\newblock Ladi-vton: Latent diffusion textual-inversion enhanced virtual try-on.
\newblock In \emph{Proceedings of the 31st ACM international conference on multimedia}, pages 8580--8589, 2023.

\bibitem[Xie et~al.(2023)Xie, Huang, Dong, Zhao, Dong, Zhang, Zhu, and Liang]{xie2023gp}
Zhenyu Xie, Zaiyu Huang, Xin Dong, Fuwei Zhao, Haoye Dong, Xijin Zhang, Feida Zhu, and Xiaodan Liang.
\newblock Gp-vton: Towards general purpose virtual try-on via collaborative local-flow global-parsing learning.
\newblock In \emph{Proceedings of the IEEE/CVF conference on computer vision and pattern recognition}, pages 23550--23559, 2023.

\bibitem[Yan et~al.(2023)Yan, Gao, Zhang, and Xie]{yan2023linking}
Keyu Yan, Tingwei Gao, Hui Zhang, and Chengjun Xie.
\newblock Linking garment with person via semantically associated landmarks for virtual try-on.
\newblock In \emph{Proceedings of the IEEE/CVF conference on computer vision and pattern recognition}, pages 17194--17204, 2023.

\bibitem[Zhu et~al.(2023)Zhu, Yang, Zhu, Reda, Chan, Saharia, Norouzi, and Kemelmacher-Shlizerman]{zhu2023tryondiffusion}
Luyang Zhu, Dawei Yang, Tyler Zhu, Fitsum Reda, William Chan, Chitwan Saharia, Mohammad Norouzi, and Ira Kemelmacher-Shlizerman.
\newblock Tryondiffusion: A tale of two unets.
\newblock In \emph{Proceedings of the IEEE/CVF Conference on Computer Vision and Pattern Recognition}, pages 4606--4615, 2023.

\bibitem[Li et~al.(2023{\natexlab{b}})Li, Zhang, and Forsyth]{10147324}
Kedan Li, Jeffrey Zhang, and David Forsyth.
\newblock Povnet: Image-based virtual try-on through accurate warping and residual.
\newblock \emph{IEEE Transactions on Pattern Analysis and Machine Intelligence}, 45\penalty0 (10):\penalty0 12222--12235, 2023{\natexlab{b}}.
\newblock \doi{10.1109/TPAMI.2023.3283302}.

\bibitem[Yang et~al.(2024{\natexlab{c}})Yang, Ding, Hong, Huang, Tao, and Xu]{yang2024texture}
Xu~Yang, Changxing Ding, Zhibin Hong, Junhao Huang, Jin Tao, and Xiangmin Xu.
\newblock Texture-preserving diffusion models for high-fidelity virtual try-on.
\newblock In \emph{Proceedings of the IEEE/CVF conference on computer vision and pattern recognition}, pages 7017--7026, 2024{\natexlab{c}}.

\bibitem[Choi et~al.(2024)Choi, Kwak, Lee, Choi, and Shin]{choi2024improving}
Yisol Choi, Sangkyung Kwak, Kyungmin Lee, Hyungwon Choi, and Jinwoo Shin.
\newblock Improving diffusion models for authentic virtual try-on in the wild.
\newblock In \emph{European Conference on Computer Vision}, pages 206--235. Springer, 2024.

\bibitem[Zhang et~al.(2024{\natexlab{a}})Zhang, Li, Chang, and Forsyth]{zhang2024acdgvtonaccuratecontaineddiffusion}
Jeffrey Zhang, Kedan Li, Shao-Yu Chang, and David Forsyth.
\newblock Acdg-vton: Accurate and contained diffusion generation for virtual try-on, 2024{\natexlab{a}}.
\newblock URL \url{https://arxiv.org/abs/2403.13951}.

\bibitem[Kim et~al.(2024{\natexlab{b}})Kim, Gu, Park, Park, and Choo]{kim2024stableviton}
Jeongho Kim, Guojung Gu, Minho Park, Sunghyun Park, and Jaegul Choo.
\newblock Stableviton: Learning semantic correspondence with latent diffusion model for virtual try-on.
\newblock In \emph{Proceedings of the IEEE/CVF conference on computer vision and pattern recognition}, pages 8176--8185, 2024{\natexlab{b}}.

\bibitem[Li et~al.(2024)Li, Zhang, Chang, and Forsyth]{li2024controlling}
Kedan Li, Jeffrey Zhang, Shao-Yu Chang, and David Forsyth.
\newblock Controlling virtual try-on pipeline through rendering policies.
\newblock In \emph{Proceedings of the IEEE/CVF winter conference on applications of computer vision}, pages 5866--5875, 2024.

\bibitem[Xu et~al.(2025)Xu, Gu, Chen, and Chen]{xu2025ootdiffusion}
Yuhao Xu, Tao Gu, Weifeng Chen, and Arlene Chen.
\newblock Ootdiffusion: Outfitting fusion based latent diffusion for controllable virtual try-on.
\newblock In \emph{Proceedings of the AAAI Conference on Artificial Intelligence}, volume~39, pages 8996--9004, 2025.

\bibitem[Chong et~al.()Chong, Dong, Li, Zhang, Zhao, Jiang, Liang, et~al.]{chongcatvton}
Zheng Chong, Xiao Dong, Haoxiang Li, Wenqing Zhang, Hanqing Zhao, Dongmei Jiang, Xiaodan Liang, et~al.
\newblock Catvton: Concatenation is all you need for virtual try-on with diffusion models.
\newblock In \emph{The Thirteenth International Conference on Learning Representations}.

\bibitem[Goodfellow et~al.(2014)Goodfellow, Pouget-Abadie, Mirza, Xu, Warde-Farley, Ozair, Courville, and Bengio]{goodfellow2014generative}
Ian~J Goodfellow, Jean Pouget-Abadie, Mehdi Mirza, Bing Xu, David Warde-Farley, Sherjil Ozair, Aaron Courville, and Yoshua Bengio.
\newblock Generative adversarial nets.
\newblock \emph{Advances in neural information processing systems}, 27, 2014.

\bibitem[Rombach et~al.(2022)Rombach, Blattmann, Lorenz, Esser, and Ommer]{rombach2022high}
Robin Rombach, Andreas Blattmann, Dominik Lorenz, Patrick Esser, and Bj{\"o}rn Ommer.
\newblock High-resolution image synthesis with latent diffusion models.
\newblock In \emph{Proceedings of the IEEE/CVF conference on computer vision and pattern recognition}, pages 10684--10695, 2022.

\bibitem[Podell et~al.(2023)Podell, English, Lacey, Blattmann, Dockhorn, M{\"u}ller, Penna, and Rombach]{podell2023sdxl}
Dustin Podell, Zion English, Kyle Lacey, Andreas Blattmann, Tim Dockhorn, Jonas M{\"u}ller, Joe Penna, and Robin Rombach.
\newblock Sdxl: Improving latent diffusion models for high-resolution image synthesis.
\newblock \emph{arXiv preprint arXiv:2307.01952}, 2023.

\bibitem[Peebles and Xie(2023)]{peebles2023scalable}
William Peebles and Saining Xie.
\newblock Scalable diffusion models with transformers.
\newblock In \emph{Proceedings of the IEEE/CVF international conference on computer vision}, pages 4195--4205, 2023.

\bibitem[Zhu et~al.(2024{\natexlab{a}})Zhu, Li, Liu, Peng, Yang, and Kemelmacher-Shlizerman]{zhu2024m}
Luyang Zhu, Yingwei Li, Nan Liu, Hao Peng, Dawei Yang, and Ira Kemelmacher-Shlizerman.
\newblock M\&m vto: Multi-garment virtual try-on and editing.
\newblock In \emph{Proceedings of the IEEE/CVF Conference on Computer Vision and Pattern Recognition}, pages 1346--1356, 2024{\natexlab{a}}.

\bibitem[Ma et~al.(2017)Ma, Jia, Sun, Schiele, Tuytelaars, and Van~Gool]{ma2017pose}
Liqian Ma, Xu~Jia, Qianru Sun, Bernt Schiele, Tinne Tuytelaars, and Luc Van~Gool.
\newblock Pose guided person image generation.
\newblock \emph{Advances in neural information processing systems}, 30, 2017.

\bibitem[Balakrishnan et~al.(2018)Balakrishnan, Zhao, Dalca, Durand, and Guttag]{balakrishnan2018synthesizing}
Guha Balakrishnan, Amy Zhao, Adrian~V Dalca, Fredo Durand, and John Guttag.
\newblock Synthesizing images of humans in unseen poses.
\newblock In \emph{Proceedings of the IEEE conference on computer vision and pattern recognition}, pages 8340--8348, 2018.

\bibitem[Esser et~al.(2018)Esser, Sutter, and Ommer]{esser2018variational}
Patrick Esser, Ekaterina Sutter, and Bj{\"o}rn Ommer.
\newblock A variational u-net for conditional appearance and shape generation.
\newblock In \emph{Proceedings of the IEEE conference on computer vision and pattern recognition}, pages 8857--8866, 2018.

\bibitem[Siarohin et~al.(2018)Siarohin, Sangineto, Lathuiliere, and Sebe]{siarohin2018deformable}
Aliaksandr Siarohin, Enver Sangineto, St{\'e}phane Lathuiliere, and Nicu Sebe.
\newblock Deformable gans for pose-based human image generation.
\newblock In \emph{Proceedings of the IEEE conference on computer vision and pattern recognition}, pages 3408--3416, 2018.

\bibitem[Men et~al.(2020)Men, Mao, Jiang, Ma, and Lian]{men2020controllable}
Yifang Men, Yiming Mao, Yuning Jiang, Wei-Ying Ma, and Zhouhui Lian.
\newblock Controllable person image synthesis with attribute-decomposed gan.
\newblock In \emph{Proceedings of the IEEE/CVF conference on computer vision and pattern recognition}, pages 5084--5093, 2020.

\bibitem[Tang et~al.(2020)Tang, Bai, Zhang, Torr, and Sebe]{tang2020xinggan}
Hao Tang, Song Bai, Li~Zhang, Philip~HS Torr, and Nicu Sebe.
\newblock Xinggan for person image generation.
\newblock In \emph{Computer Vision--ECCV 2020: 16th European Conference, Glasgow, UK, August 23--28, 2020, Proceedings, Part XXV 16}, pages 717--734. Springer, 2020.

\bibitem[Raj et~al.(2018)Raj, Sangkloy, Chang, Hays, Ceylan, and Lu]{raj2018swapnet}
Amit Raj, Patsorn Sangkloy, Huiwen Chang, James Hays, Duygu Ceylan, and Jingwan Lu.
\newblock Swapnet: Image based garment transfer.
\newblock In \emph{Computer Vision--ECCV 2018: 15th European Conference, Munich, Germany, September 8--14, 2018, Proceedings, Part XII 15}, pages 679--695. Springer, 2018.

\bibitem[Han et~al.(2019)Han, Hu, Huang, and Scott]{han2019clothflow}
Xintong Han, Xiaojun Hu, Weilin Huang, and Matthew~R Scott.
\newblock Clothflow: A flow-based model for clothed person generation.
\newblock In \emph{Proceedings of the IEEE/CVF international conference on computer vision}, pages 10471--10480, 2019.

\bibitem[Sarkar et~al.(2020)Sarkar, Mehta, Xu, Golyanik, and Theobalt]{sarkar2020neural}
Kripasindhu Sarkar, Dushyant Mehta, Weipeng Xu, Vladislav Golyanik, and Christian Theobalt.
\newblock Neural re-rendering of humans from a single image.
\newblock In \emph{Computer Vision--ECCV 2020: 16th European Conference, Glasgow, UK, August 23--28, 2020, Proceedings, Part XI 16}, pages 596--613. Springer, 2020.

\bibitem[Cui et~al.(2021)Cui, McKee, and Lazebnik]{Cui_2021_ICCV}
Aiyu Cui, Daniel McKee, and Svetlana Lazebnik.
\newblock Dressing in order: Recurrent person image generation for pose transfer, virtual try-on and outfit editing.
\newblock In \emph{Proceedings of the IEEE/CVF International Conference on Computer Vision (ICCV)}, pages 14638--14647, October 2021.

\bibitem[Sarkar et~al.(2021)Sarkar, Golyanik, Liu, and Theobalt]{sarkar2021style}
Kripasindhu Sarkar, Vladislav Golyanik, Lingjie Liu, and Christian Theobalt.
\newblock Style and pose control for image synthesis of humans from a single monocular view.
\newblock \emph{arXiv preprint arXiv:2102.11263}, 2021.

\bibitem[Kuppa et~al.(2021)Kuppa, Jong, Liu, Liu, and Moh]{kuppa2021shineon}
Gaurav Kuppa, Andrew Jong, Xin Liu, Ziwei Liu, and Teng-Sheng Moh.
\newblock Shineon: Illuminating design choices for practical video-based virtual clothing try-on.
\newblock In \emph{Proceedings of the IEEE/CVF Winter Conference on Applications of Computer Vision}, pages 191--200, 2021.

\bibitem[Zhong et~al.(2021)Zhong, Wu, Tan, Lin, and Wu]{zhong2021mv}
Xiaojing Zhong, Zhonghua Wu, Taizhe Tan, Guosheng Lin, and Qingyao Wu.
\newblock Mv-ton: Memory-based video virtual try-on network.
\newblock In \emph{Proceedings of the 29th ACM International Conference on Multimedia}, pages 908--916, 2021.

\bibitem[Jiang et~al.(2022)Jiang, Wang, Yan, and Liu]{jiang2022clothformer}
Jianbin Jiang, Tan Wang, He~Yan, and Junhui Liu.
\newblock Clothformer: Taming video virtual try-on in all module.
\newblock In \emph{Proceedings of the IEEE/CVF Conference on Computer Vision and Pattern Recognition}, pages 10799--10808, 2022.

\bibitem[Xu et~al.(2024{\natexlab{b}})Xu, Chen, Wang, Xing, Zhai, Sang, Lan, Xiao, and Gao]{xu2024tunnel}
Zhengze Xu, Mengting Chen, Zhao Wang, Linyu Xing, Zhonghua Zhai, Nong Sang, Jinsong Lan, Shuai Xiao, and Changxin Gao.
\newblock Tunnel try-on: Excavating spatial-temporal tunnels for high-quality virtual try-on in videos.
\newblock In \emph{Proceedings of the 32nd ACM International Conference on Multimedia}, pages 3199--3208, 2024{\natexlab{b}}.

\bibitem[Fang et~al.(2024)Fang, Zhai, Su, Song, Zhu, Wang, Chen, Liu, Cao, and Zha]{fang2024vivid}
Zixun Fang, Wei Zhai, Aimin Su, Hongliang Song, Kai Zhu, Mao Wang, Yu~Chen, Zhiheng Liu, Yang Cao, and Zheng-Jun Zha.
\newblock Vivid: Video virtual try-on using diffusion models.
\newblock \emph{arXiv preprint arXiv:2405.11794}, 2024.

\bibitem[Wang et~al.(2024{\natexlab{b}})Wang, Dai, Chan, Zhou, Zhang, and Liu]{wang2024gpd}
Yuanbin Wang, Weilun Dai, Long Chan, Huanyu Zhou, Aixi Zhang, and Si~Liu.
\newblock Gpd-vvto: Preserving garment details in video virtual try-on.
\newblock In \emph{Proceedings of the 32nd ACM International Conference on Multimedia}, pages 7133--7142, 2024{\natexlab{b}}.

\bibitem[He et~al.(2024)He, Chen, Wang, Li, Torr, and Lin]{he2024wildvidfit}
Zijian He, Peixin Chen, Guangrun Wang, Guanbin Li, Philip~HS Torr, and Liang Lin.
\newblock Wildvidfit: Video virtual try-on in the wild via image-based controlled diffusion models.
\newblock In \emph{European Conference on Computer Vision}, pages 123--139. Springer, 2024.

\bibitem[Chan et~al.(2019)Chan, Ginosar, Zhou, and Efros]{chan2019dance}
Caroline Chan, Shiry Ginosar, Tinghui Zhou, and Alexei~A Efros.
\newblock Everybody dance now.
\newblock In \emph{IEEE International Conference on Computer Vision (ICCV)}, 2019.

\bibitem[Siarohin et~al.(2019)Siarohin, Lathuilière, Tulyakov, Ricci, and Sebe]{Siarohin_2019_NeurIPS}
Aliaksandr Siarohin, Stéphane Lathuilière, Sergey Tulyakov, Elisa Ricci, and Nicu Sebe.
\newblock First order motion model for image animation.
\newblock In \emph{Conference on Neural Information Processing Systems (NeurIPS)}, December 2019.

\bibitem[Holynski et~al.(2021)Holynski, Curless, Seitz, and Szeliski]{Holynski_2021_CVPR}
Aleksander Holynski, Brian~L. Curless, Steven~M. Seitz, and Richard Szeliski.
\newblock Animating pictures with eulerian motion fields.
\newblock In \emph{Proceedings of the IEEE/CVF Conference on Computer Vision and Pattern Recognition (CVPR)}, pages 5810--5819, June 2021.

\bibitem[Siarohin et~al.(2021)Siarohin, Woodford, Ren, Chai, and Tulyakov]{siarohin2021motion}
Aliaksandr Siarohin, Oliver Woodford, Jian Ren, Menglei Chai, and Sergey Tulyakov.
\newblock Motion representations for articulated animation.
\newblock In \emph{CVPR}, 2021.

\bibitem[Xu et~al.(2024{\natexlab{c}})Xu, Zhang, Liew, Yan, Liu, Zhang, Feng, and Shou]{xu2023magicanimate}
Zhongcong Xu, Jianfeng Zhang, Jun~Hao Liew, Hanshu Yan, Jia-Wei Liu, Chenxu Zhang, Jiashi Feng, and Mike~Zheng Shou.
\newblock Magicanimate: Temporally consistent human image animation using diffusion model.
\newblock 2024{\natexlab{c}}.

\bibitem[Guo et~al.(2024)Guo, Yang, Rao, Liang, Wang, Qiao, Agrawala, Lin, and Dai]{guo2024animatediff}
Yuwei Guo, Ceyuan Yang, Anyi Rao, Zhengyang Liang, Yaohui Wang, Yu~Qiao, Maneesh Agrawala, Dahua Lin, and Bo~Dai.
\newblock Animatediff: Animate your personalized text-to-image diffusion models without specific tuning.
\newblock In \emph{The Twelfth International Conference on Learning Representations}, 2024.
\newblock URL \url{https://openreview.net/forum?id=Fx2SbBgcte}.

\bibitem[Karras et~al.(2023)Karras, Holynski, Wang, and Kemelmacher-Shlizerman]{dreampose_2023}
Johanna Karras, Aleksander Holynski, Ting-Chun Wang, and Ira Kemelmacher-Shlizerman.
\newblock Dreampose: Fashion image-to-video synthesis via stable diffusion.
\newblock 2023.

\bibitem[Zhang et~al.(2024{\natexlab{b}})Zhang, Gu, Wang, Wang, Cheng, Zhu, and Zou]{mimicmotion2024}
Yuang Zhang, Jiaxi Gu, Li-Wen Wang, Han Wang, Junqi Cheng, Yuefeng Zhu, and Fangyuan Zou.
\newblock Mimicmotion: High-quality human motion video generation with confidence-aware pose guidance.
\newblock \emph{arXiv preprint arXiv:2406.19680}, 2024{\natexlab{b}}.

\bibitem[Azadi et~al.(2023)Azadi, Shah, Hayes, Parikh, and Gupta]{azadi2023make}
Samaneh Azadi, Akbar Shah, Thomas Hayes, Devi Parikh, and Sonal Gupta.
\newblock Make-an-animation: Large-scale text-conditional 3d human motion generation.
\newblock In \emph{Proceedings of the IEEE/CVF International Conference on Computer Vision}, pages 15039--15048, 2023.

\bibitem[Tevet et~al.(2022)Tevet, Raab, Gordon, Shafir, Cohen-Or, and Bermano]{tevet2022human}
Guy Tevet, Sigal Raab, Brian Gordon, Yonatan Shafir, Daniel Cohen-Or, and Amit~H Bermano.
\newblock Human motion diffusion model.
\newblock \emph{arXiv preprint arXiv:2209.14916}, 2022.

\bibitem[Guo et~al.(2022)Guo, Zou, Zuo, Wang, Ji, Li, and Cheng]{guo2022generating}
Chuan Guo, Shihao Zou, Xinxin Zuo, Sen Wang, Wei Ji, Xingyu Li, and Li~Cheng.
\newblock Generating diverse and natural 3d human motions from text.
\newblock In \emph{Proceedings of the IEEE/CVF conference on computer vision and pattern recognition}, pages 5152--5161, 2022.

\bibitem[Hu(2024)]{hu2024animate}
Li~Hu.
\newblock Animate anyone: Consistent and controllable image-to-video synthesis for character animation.
\newblock In \emph{Proceedings of the IEEE/CVF Conference on Computer Vision and Pattern Recognition}, pages 8153--8163, 2024.

\bibitem[Zhu et~al.(2024{\natexlab{b}})Zhu, Chen, Dai, Dong, Xu, Cao, Yao, Zhu, and Zhu]{zhu2024champ}
Shenhao Zhu, Junming~Leo Chen, Zuozhuo Dai, Zilong Dong, Yinghui Xu, Xun Cao, Yao Yao, Hao Zhu, and Siyu Zhu.
\newblock Champ: Controllable and consistent human image animation with 3d parametric guidance.
\newblock In \emph{European Conference on Computer Vision}, pages 145--162. Springer, 2024{\natexlab{b}}.

\bibitem[Dong et~al.(2019)Dong, Liang, Shen, Wu, Chen, and Yin]{dong2019fw}
Haoye Dong, Xiaodan Liang, Xiaohui Shen, Bowen Wu, Bing-Cheng Chen, and Jian Yin.
\newblock Fw-gan: Flow-navigated warping gan for video virtual try-on.
\newblock In \emph{Proceedings of the IEEE/CVF international conference on computer vision}, pages 1161--1170, 2019.

\bibitem[Weng et~al.(2022)Weng, Curless, Srinivasan, Barron, and Kemelmacher-Shlizerman]{weng_humannerf_2022_cvpr}
Chung-Yi Weng, Brian Curless, Pratul~P. Srinivasan, Jonathan~T. Barron, and Ira Kemelmacher-Shlizerman.
\newblock Human{N}e{RF}: Free-viewpoint rendering of moving people from monocular video.
\newblock In \emph{Proceedings of the IEEE/CVF Conference on Computer Vision and Pattern Recognition (CVPR)}, pages 16210--16220, June 2022.

\bibitem[Liu et~al.(2021)Liu, Habermann, Rudnev, Sarkar, Gu, and Theobalt]{liu2021neural}
Lingjie Liu, Marc Habermann, Viktor Rudnev, Kripasindhu Sarkar, Jiatao Gu, and Christian Theobalt.
\newblock Neural actor: Neural free-view synthesis of human actors with pose control.
\newblock \emph{ACM Trans. Graph.(ACM SIGGRAPH Asia)}, 2021.

\bibitem[AI(2024)]{kling}
Kling AI.
\newblock Kling ai: Next-generation ai creative studio, 2024.
\newblock URL \url{https://klingai.com/}.

\bibitem[Couairon et~al.(2022)Couairon, Verbeek, Schwenk, and Cord]{diffedit}
Guillaume Couairon, Jakob Verbeek, Holger Schwenk, and Matthieu Cord.
\newblock Diffedit: Diffusion-based semantic image editing with mask guidance.
\newblock \emph{ArXiv}, abs/2210.11427, 2022.
\newblock URL \url{https://api.semanticscholar.org/CorpusID:253018768}.

\bibitem[OpenAI(2023)]{gpt}
OpenAI.
\newblock {GPT-4} technical report.
\newblock \emph{arXiv preprint arXiv:2303.08774}, 2023.

\bibitem[Huang et~al.(2024{\natexlab{a}})Huang, He, Yu, Zhang, Si, Jiang, Zhang, Wu, Jin, Chanpaisit, Wang, Chen, Wang, Lin, Qiao, and Liu]{vbench}
Ziqi Huang, Yinan He, Jiashuo Yu, Fan Zhang, Chenyang Si, Yuming Jiang, Yuanhan Zhang, Tianxing Wu, Qingyang Jin, Nattapol Chanpaisit, Yaohui Wang, Xinyuan Chen, Limin Wang, Dahua Lin, Yu~Qiao, and Ziwei Liu.
\newblock {VBench}: Comprehensive benchmark suite for video generative models.
\newblock In \emph{Proceedings of the IEEE/CVF Conference on Computer Vision and Pattern Recognition}, 2024{\natexlab{a}}.

\bibitem[Huang et~al.(2024{\natexlab{b}})Huang, Zhang, Xu, He, Yu, Dong, Ma, Chanpaisit, Si, Jiang, Wang, Chen, Chen, Wang, Lin, Qiao, and Liu]{vbench++}
Ziqi Huang, Fan Zhang, Xiaojie Xu, Yinan He, Jiashuo Yu, Ziyue Dong, Qianli Ma, Nattapol Chanpaisit, Chenyang Si, Yuming Jiang, Yaohui Wang, Xinyuan Chen, Ying-Cong Chen, Limin Wang, Dahua Lin, Yu~Qiao, and Ziwei Liu.
\newblock {VBench++}: Comprehensive and versatile benchmark suite for video generative models.
\newblock \emph{arXiv preprint arXiv:2411.13503}, 2024{\natexlab{b}}.

\bibitem[Zheng et~al.(2025)Zheng, Huang, Liu, Zou, He, Zhang, Zhang, He, Zheng, Qiao, and Liu]{vbench2}
Dian Zheng, Ziqi Huang, Hongbo Liu, Kai Zou, Yinan He, Fan Zhang, Yuanhan Zhang, Jingwen He, Wei-Shi Zheng, Yu~Qiao, and Ziwei Liu.
\newblock {VBench-2.0}: Advancing video generation benchmark suite for intrinsic faithfulness.
\newblock \emph{arXiv preprint arXiv:2503.21755}, 2025.

\end{thebibliography}
}



\appendix




\end{document}